\pdfoutput=1

\documentclass[11pt, a4paper]{article}

\usepackage[]{naacl2021}

\usepackage{times}
\usepackage{latexsym}
\usepackage{graphicx}
\usepackage{amsmath, amssymb}
\usepackage{caption}

\usepackage{placeins} 
\usepackage{stmaryrd}
\usepackage{url}
\usepackage{xr}
\usepackage[T1]{fontenc}

\usepackage[utf8]{inputenc}

\usepackage{microtype}
\usepackage{tikz}
\usetikzlibrary{positioning}

\graphicspath{{graphs/}}
\DeclareMathOperator{\EX}{\mathbb{E}}
\newcommand{\el}{\EX_{layer}}

\usepackage{xr}

\makeatletter
\newcommand*{\addFileDependency}[1]{
  \typeout{(#1)}
  \@addtofilelist{#1}
  \IfFileExists{#1}{}{\typeout{No file #1.}}
}
\makeatother

\newcommand*{\myexternaldocument}[1]{%
    \externaldocument{#1}%
    \addFileDependency{#1.tex}%
    \addFileDependency{#1.aux}%
}
\myexternaldocument{appendix}

\title{Mediators in Determining what Processing BERT Performs First}
\author{Aviv Slobodkin \hspace{1.4cm} Leshem Choshen \hspace{1.4cm} Omri Abend \\
        School of Computer Science and Engineering\\
        The Hebrew University of Jerusalem\\
        {\tt \{aviv.slobodkin,leshem.choshen,omri.abend\}@mail.huji.ac.il}}

\date{}

\begin{document}
\maketitle
\begin{abstract}
    Probing neural models for the ability to perform downstream tasks using their activation patterns is often used to localize what parts of the network specialize in performing what tasks. 
    However, little work addressed potential mediating factors in such comparisons. As a test-case mediating factor, we consider the prediction's {\it context length}, namely the length of the span whose processing is minimally required to perform the prediction.
    We show that not controlling for context length may lead to contradictory conclusions as to the localization patterns of the network, depending on the distribution of the probing dataset. 
    Indeed, when probing BERT with seven tasks, we find that it is possible to get 196 different rankings between them when manipulating the distribution of context lengths in the probing dataset. 
    We conclude by presenting best practices for conducting such comparisons in the future.\footnote{The code is available at \url{https://github.com/lovodkin93/BERT-context-distance}.}
\end{abstract}

\section{Introduction}

The strong performance of end-to-end models and the difficulty in understanding their inner workings has led to extensive research aimed at interpreting their behavior \citep{li-etal-2016-visualizing, DBLP:journals/corr/YosinskiCNFL15, karpathy2015visualizing}.
This notion has led researchers to investigate the behavioral traits of networks in general \citep{Li2015ConvergentLD, hacohen2019all} and representative architectures in particular  \citep{Schlichtkrull2020InterpretingGN}. Within NLP, Transformer-based pretrained embeddings are the basis for many tasks, which underscores the importance in interpreting their behavior \citep{belinkov-etal-2020-linguistic}, and especially the behavior of BERT \cite{devlin-etal-2019-bert, rogers2020primer}, perhaps the most widely used of Transformer-based models.

In this work, we analyze the common approach of {\it probing} (\S\ref{subsec:Edge-probing}), used to localize where ``knowledge'' of particular tasks is encoded; localization is often carried out in terms of the layers most responsible for the task at hand \citep[c.f.][]{tenney2018what}.
Various works \cite{tenney-etal-2019-bert, peters-etal-2018-dissecting, blevins-etal-2018-deep} showed that some tasks are processed in lower levels than others. 

We examine the extent to which potential mediating factors may account for observed trends and show that varying some mediating factors (see \S\ref{subsec:Mediation Analysis}) may diminish, or even reverse, the conclusions made by \citet[T19;][]{tenney-etal-2019-bert}.
Specifically, despite reaffirming T19's experimental findings, we contest T19's interpretation of the results, namely that the processing carried out by BERT parallels the classical NLP pipeline. Indeed, T19 concludes that lexical tasks (POS tagging) are performed by the lower layers, followed by syntactic tasks, whereas more semantic tasks are performed later on. This analysis rests on the assumption that the nature of the task (lexical, syntactic, or semantic) is the driving force that determines what layer performs what analysis. We show that other factors should be weighed in as well. Specifically, we show that manipulating the distribution of examples in the probing dataset can lead to a variety of different conclusions as to what tasks are performed first.

We argue that potential mediators must be considered when comparing tasks, and focus on one such mediator -- the \emph{context length}, which we define as the number of tokens whose processing is minimally required to perform the prediction. We operationalize this notion by defining it as the maximal distance between any two tokens for which a label is predicted. This amounts to the span length in tasks that involve a single span (e.g., NER), and to the dependency length in tasks that address the relation between two spans. See \S\ref{subsec:Mediation Analysis}.
Our motivation for considering context length as a mediator is grounded in previous work that presented the difficulty posed by long-distance dependencies in various NLP tasks \citep{10.5555/1620754.1620790, sennrich-2017-grammatical}, and particularly in previous work that indicated the Transformers' difficulty to generalize across different dependency lengths \citep{choshen-abend-2019-automatically}.

We show that in some of the cases where one task seems to be better predicted by a higher layer than another task, controlling for context length may reverse that order. Indeed we show that 196 different rankings between the seven tasks explored in T19 may be obtained with a suitable distribution over the probing datasets, namely 196 different ways to rank the tasks according to their expected layer. Moreover, our results show that when context length is not taken into account, one task (e.g., dependency parsing) may seem to be processed at a higher layer than another (e.g., NER), when its expected layer (see \S\ref{subsec:Expected Layer}) is, in fact, lower for all ranges of context lengths (\S\ref{subsubsec:Interesting Case}).


\section{Background}\label{sec:background}

We begin by laying out the terminology and methodology we will use in the paper.

\paragraph{Edge Probing.}\label{subsec:Edge-probing}
Edge probing is the method of training a classifier for a given task on different parts of the network (without fine-tuning). Success in classification is interpreted as evidence that the required features for classification are somehow encoded in the examined part and are sufficiently easy to extract. In our experiments, we follow T19 and probe BERT with Named Entity Recognition (NER), a constituent-based task (classifying Non-terminals - Non-term.), Semantic Role Labeling (SRL), Co-reference (Co-ref.), Semantic Proto-Roles \citep[SPR;][]{reisinger-etal-2015-semantic}, Relation Classification (RC) and the Stanford Dependency Parsing \citep[Dep.;][]{de-marneffe-etal-2006-generating}.

Causal considerations in interpreting probing results were also emphasized by several recent works \citep[e.g.,][]{Kaushik2020Learning,vig2020causal,elazar2021amnesic}.

\paragraph{Localization by Expected Layer.}\label{subsec:Expected Layer}
The expected layer metric (which we will henceforth refer to it as $\el$) of T19 assesses which layer in BERT is most needed for prediction: a probing classifier $P^{(l)}$ is trained on the lowest $l$ layers.
Then, a differential score $\Delta^{(l)}$ is computed, which indicates the performance gain when taking into account one additional layer:

\begin{equation} \label{eq1}
    \Delta^{(l)} = Score(P^{(l)})-Score(P^{(l-1)})
\end{equation}
Once all the $\{\Delta^{(l)}\}^{12}_{l=1}$ are computed, we may compute $\el$:
\begin{equation} \label{eq2}
     \el[l]=\frac{\sum_{l=1}^{12}{l\cdot\Delta^{(l)}}}{\sum_{l=1}^{12}{\Delta^{(l)}}}
\end{equation}

Therefore, unlike standard edge probing, which is performed on each layer individually, computing $\el$ takes into account all layers up to a given $l$.

\paragraph{Mediation Analysis.} \label{subsec:Mediation Analysis}
Each of the explored tasks classifies one or two input sub-spans. In both cases, we define the context length to be the distance between the earliest and latest span index. Namely, for tasks with two spans (e.g., SPR), $span_1$=[$i_1$,$j_1$] and $span_2$=[$i_2$,$j_2$], where $span_1$ appears before $span_2$, the context length is $j_2$-$i_1$, whereas for tasks with just one span (e.g., NER), $span_1$=[$i_1$,$j_1$], it is $j_1$-$i_1$.

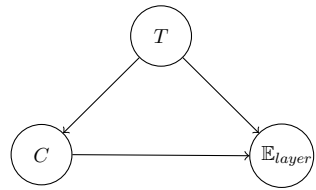
\begin{figure}[htp]
\centering
\begin{tikzpicture}[roundnode/.style={circle, draw,text width=0.3in,align=center, minimum size=1mm}]
\node[roundnode, scale=0.7]      (Task)                              {$T$};
\node[roundnode, scale=0.7]        (Length)       [below left=of Task] {$C$};
\node[roundnode, scale=0.7]      (Expected_Layer)       [below right=of Task] {$\el$};

\draw[->] (Task.south west) -- (Length.north east);
\draw[->] (Task.south east) -- (Expected_Layer.north west);
\draw[->] (Length.east) -- (Expected_Layer.west);
\end{tikzpicture}
\caption{The relationship we stipulate between the task, the context length, and $\el$. We use two random variables: $T$ is the task, which can be any of the seven tasks we observe and $C$ is the context length.}
\label{fig:mediator}
\end{figure}

In order to examine the effect of context length on $\el$, we model it as a mediating factor, namely as an intermediate variable that (partly) explains the relationship between two other variables (in this work, a task and its $\el$). See Figure \ref{fig:mediator}.

We bin each task's test set into non-overlapping bins, according to their context length ranges. We use the notation `i-j' to denote the bin of context lengths in the range [i,j]. For example, the second bin would be '3-5', denoting context lengths 3, 4, and 5. In addition, given a specific task, two possible approaches exist to examine the mediation effect of context length on the task's $\el$. The first one bins all the task's data into sub-sets, in advance. Then, this approach fine-tunes over each subset separately. Alternatively, the second approach fine-tunes over the whole dataset, binning only during the test phase.
We follow the latter approach, as it is more computationally efficient.

Interestingly, in \S\ref{subsubsec:Interesting Case}, we encounter a special edge case, where the aggregated average (i.e., $\el$) of one task is higher than another, whereas in each sub-set (by a given context length) it is lower. This may occur when the weight of the sub-sets differs between the two aggregations.

\section{Experiments} \label{sec:Experiments}
We hypothesize that the context length is a mediating factor in the $\el$ of a task. In order to test this hypothesis, we run the following experiments, aiming at isolating the context length.

We use the SPR1 dataset \citep{reisinger-etal-2015-semantic} to probe SPR, the English Web Treebank for the Dep. task \citep{silveira-etal-2014-gold}, the SemEval 2010 Task 8 for the RC task  \citep{hendrickx-etal-2009-semeval}, and the OntoNotes 5.0 dataset \citep{weischedel2013ontonotes} for the other tasks. Configurations follow the defaults in the \href{https://github.com/nyu-mll/jiant}{Jiant} toolkit implementation \citep{wang2019jiant}. In addition, we work with the BERT-base model.

\subsection{The Effect on $\el$}\label{subsec:Maximum context length}

First, we wish to confirm that context length indeed affects $\el$ and that the task is not a sole contributor to this.
Given a task and a threshold $thr$, we compile a dataset for the task containing the sub-set of examples with context lengths shorter than $thr$, and use it to compute $\el$. We do it for all tasks and for every integer threshold between 0 and a maximal threshold, which is selected separately for each task to ensure that at least 2000 instances remain in the last bin.

\begin{figure}[h!]
    \includegraphics[width=7cm]{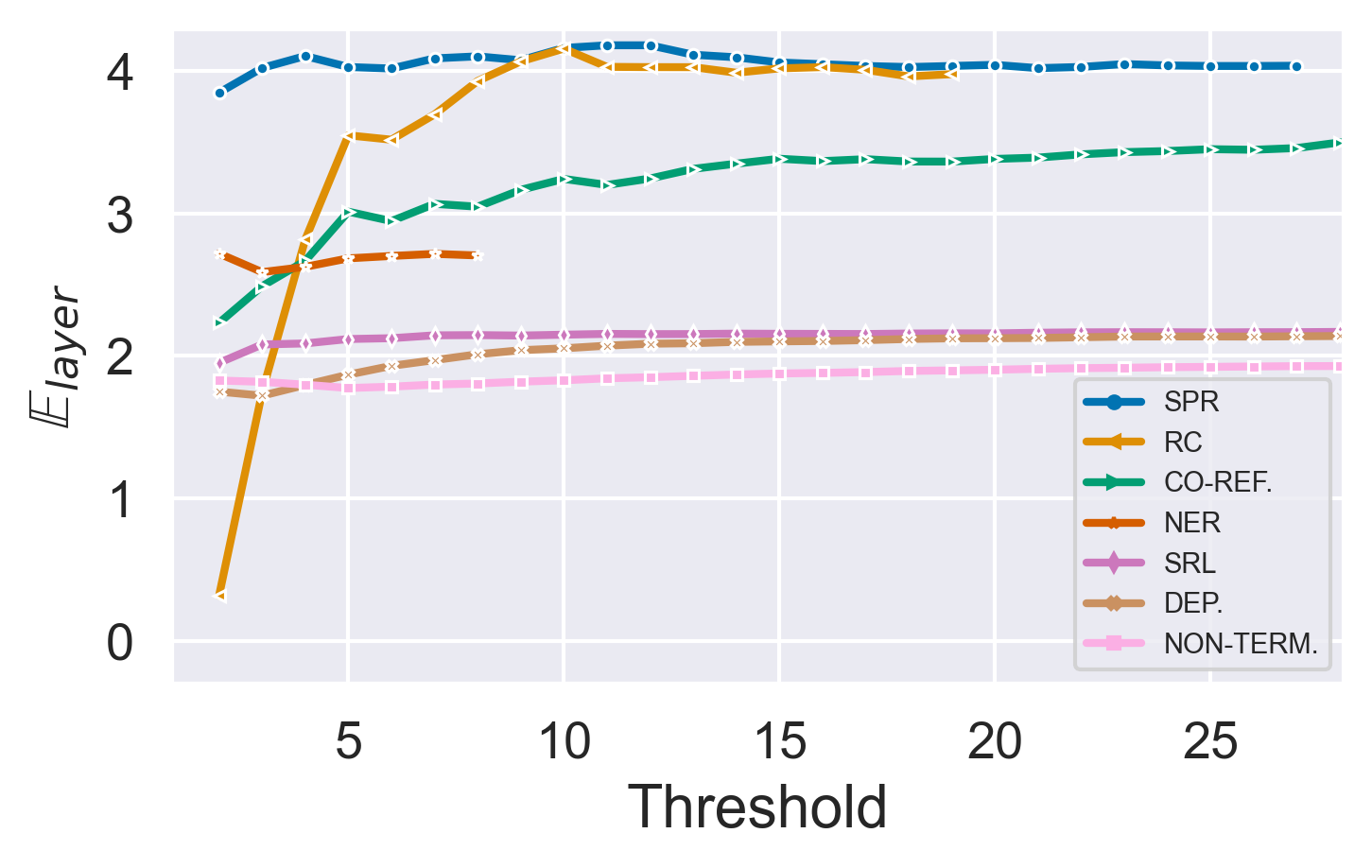}
    \captionsetup{aboveskip=0pt, belowskip=0pt}
    \caption{$\el$ as a function of a threshold on the context length. For each such threshold $thr$ (x-axis), $\el$ (y-axis) is computed based only on the examples with context length no longer than $thr$.}
    \label{fig:expected_layer_thr}
\end{figure}

We find that context length plays an important role in the difference between the expected layers (Figure \ref{fig:expected_layer_thr}). Most notably, the Co-ref., SRL, Dep., and RC tasks' $\el$ increases when increasing the threshold.

Next, we divide the data into smaller bins of non-overlapping context length ranges, in order to control for the influence of the context lengths on the expected layers of the tasks. 
We compute $\el$ for sub-sets of similar lengths. In choosing the size of each such range, we try to balance between informativeness (narrower ranges) and reliability (having enough examples in each range, so as to reduce noise). 
We find that the narrowest range width that retains at least 1\% of the examples in each bin is 3. We thus divide the dataset for each task into context length ranges of width 3, until the maximal threshold is reached. Higher context lengths are lumped into an additional bin. 

\subsubsection{Manipulating the Context Length Distribution: An Extreme Case.} \label{subsubsec:Interesting Case}

\begin{figure}[h!]
    \includegraphics[width=7cm]{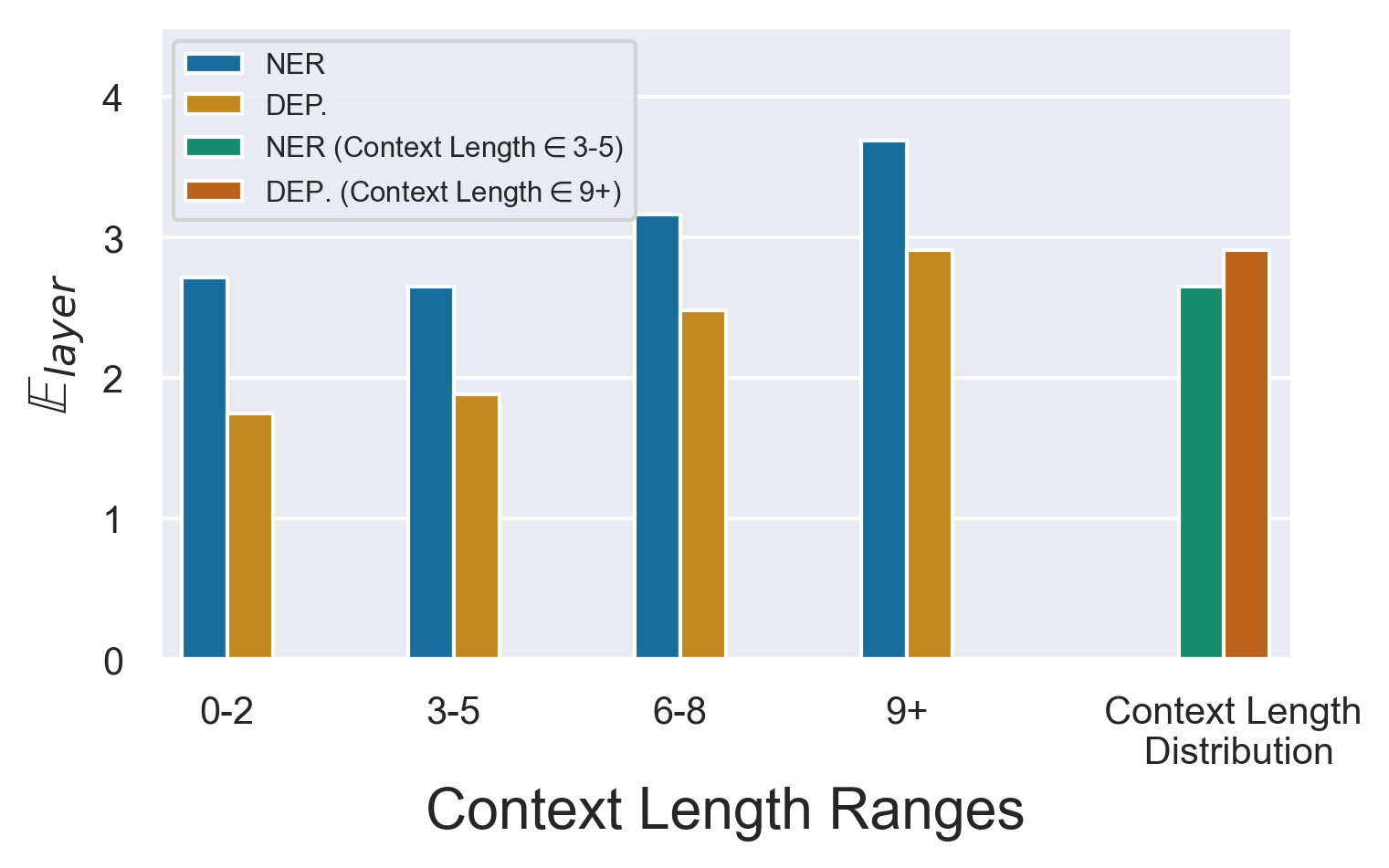}
    \captionsetup{aboveskip=0pt, belowskip=0pt}
    \caption{$\el$ of NER and Dep. for different context length ranges (4 left blue and yellow pairs), and their $\el$ when all instances of NER are of context length $l\in [3,5]$ and all those of Dep. are of context length $l\geq 9$ (rightmost green and red pair). While for every context length range, NER's $\el$ is bigger than that of Dep., for some context length distribution that order may be reversed.}
    \label{fig:interesting_case}
\end{figure}

We begin by examining two specific tasks: Dep. and NER, and their $\el$ for each context length's range. We then consider, for simplicity, a case where all the context lengths of Dep. are of length 9+, while those of NER are in the range of 3-5 (Figure \ref{fig:interesting_case}). We see that when controlling for context length, Dep. is computed in a lower layer than NER, regardless of the range. However, depending on the distribution of context lengths in the probing dataset, the outcome may be completely different, with Dep. being processed in higher layers (for a similar example of a different task-pair, see \S\ref{subsec:Additional Interesting Case}). 

These results indicate that the results of T19 do not necessarily indicate that BERT is performing a pipeline of computations (as is commonly asserted, see e.g., T19 and \citet{blevins-etal-2018-deep}), and that mediating factors need to be taken into account when interpreting $\el$.

\subsection{Imposing Similar Length Distributions} \label{subsubsec:NDE}

In the previous section, we observed that one task can be both higher and lower than another. That depends on the distribution of context lengths in the probing dataset. We next ask whether such a "paradox" arises in experiments when imposing the same context length distributions on the two tasks.

Following \citet{10.5555/647235.720084}, we employ mediation analysis and specifically concentrate on the Natural Direct Effect (NDE), which is the difference between two of the observed dependent variables (in our case $\el{}$), when fixing the mediator. In our case, the NDE is the difference between the $\el{}$ of two tasks, while forcing the same context length distribution on both. For convenience, we force the distribution of one of the examined tasks (for more details, see \S\ref{Context Length Distribution}), but any distribution is applicable. In general, the equation for computing the NDE of tasks $t_1$ and $t_2$, with the context length distribution of $t_1$ imposed on both, is:
\begin{equation} \label{eq3}
\scalebox{0.75}{$
   \begin{aligned}
        NDE_{t_1\shortrightarrow t_2} & = \sum_{c}[E_\Delta[l|C=c,T=t_2] \\
        & - E_\Delta[l|C=c,T=t_1]] \cdot P(C=c|T=t_1)  
   \end{aligned}$}
\end{equation}
where T is a random variable of the tasks, and C is a random variable of the context length.

\begin{figure}[h!]
    \includegraphics[width=7cm]{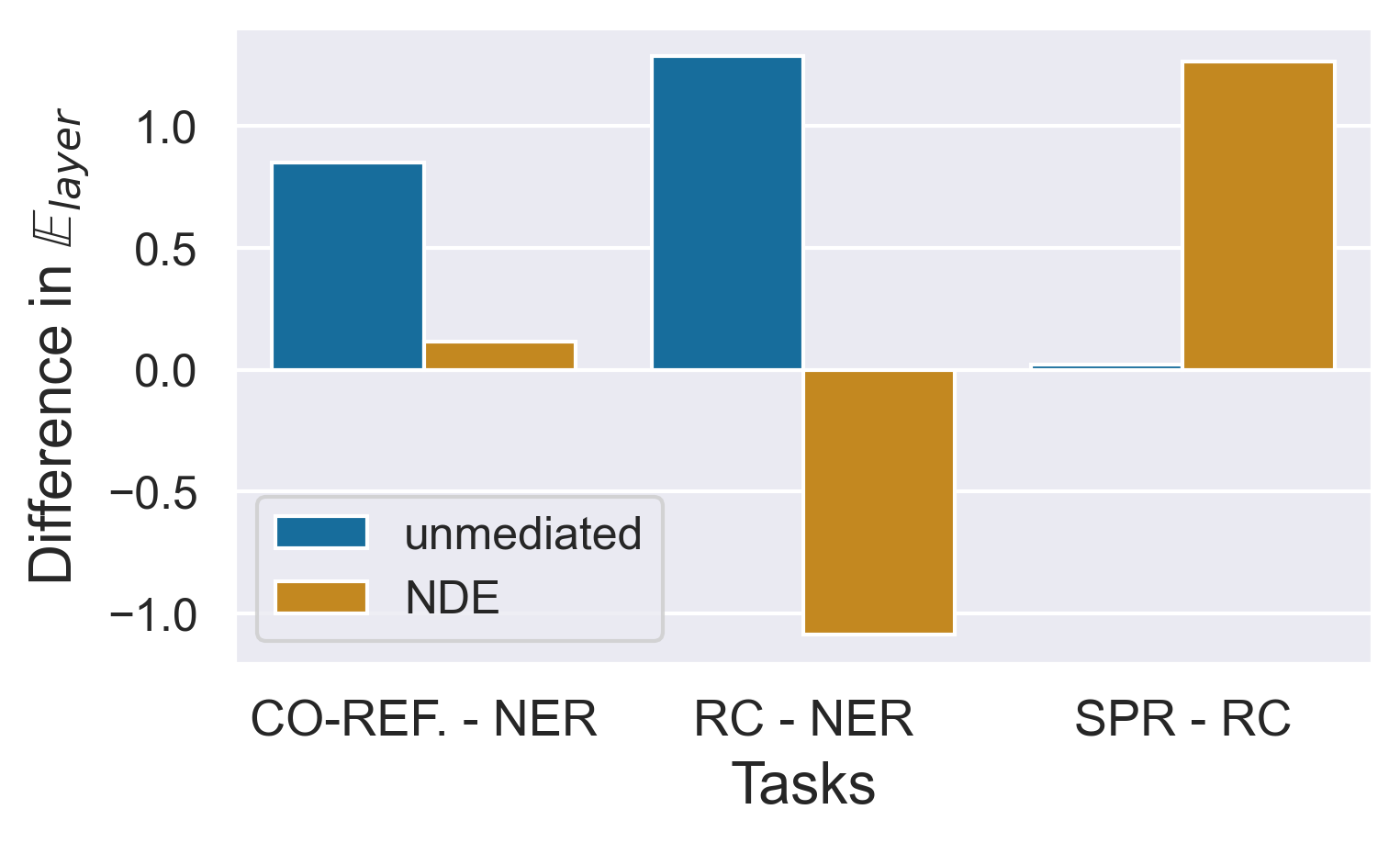}
    \captionsetup{aboveskip=0pt, belowskip=0pt}
    \caption{Difference between unmediated $\el{}$ and NDE for NER and Co-ref. (left); NER and RC (middle); and SPR and RC (right). The employed context length distributions (as part of the NDE calculations) are of Co-ref., NER and SPR, respectively.}
    \label{fig:NDE_vs_expected_layer}
\end{figure}

We apply NDE twice for every pair of tasks  (once for each task's context length distribution). We then compare the results to the difference between the tasks' expected layers where each task keeps its original context length distribution (unmediated). Results (Figure \ref{fig:NDE_vs_expected_layer}) show that the difference could be more than 50 times larger (change of 1.24 in absolute value) or decrease by 86\% (0.73 in absolute value). In some cases the order of the two tasks is reversed, namely, the task that is lower with one distribution becomes higher with another. This shows that even among our examined set of seven tasks, the effect of potential mediators cannot be ignored. For more results, see \S\ref{subsec:NDE vs. Unmediated difference for All Task-pairs}.

\subsubsection{Controlling for Context Length}  \label{subsubsec:Controlled Effect}

After observing that the distribution of context length in the probing dataset may affect the relative order of the expected layers, we propose a more detailed and accurate method to compare the expected layers, which does not rely on a specific length distribution.
We do so by plotting the {\it controlled effect},
namely $\el$ for each range separately.

\begin{figure}[h!]
    \includegraphics[width=7cm]{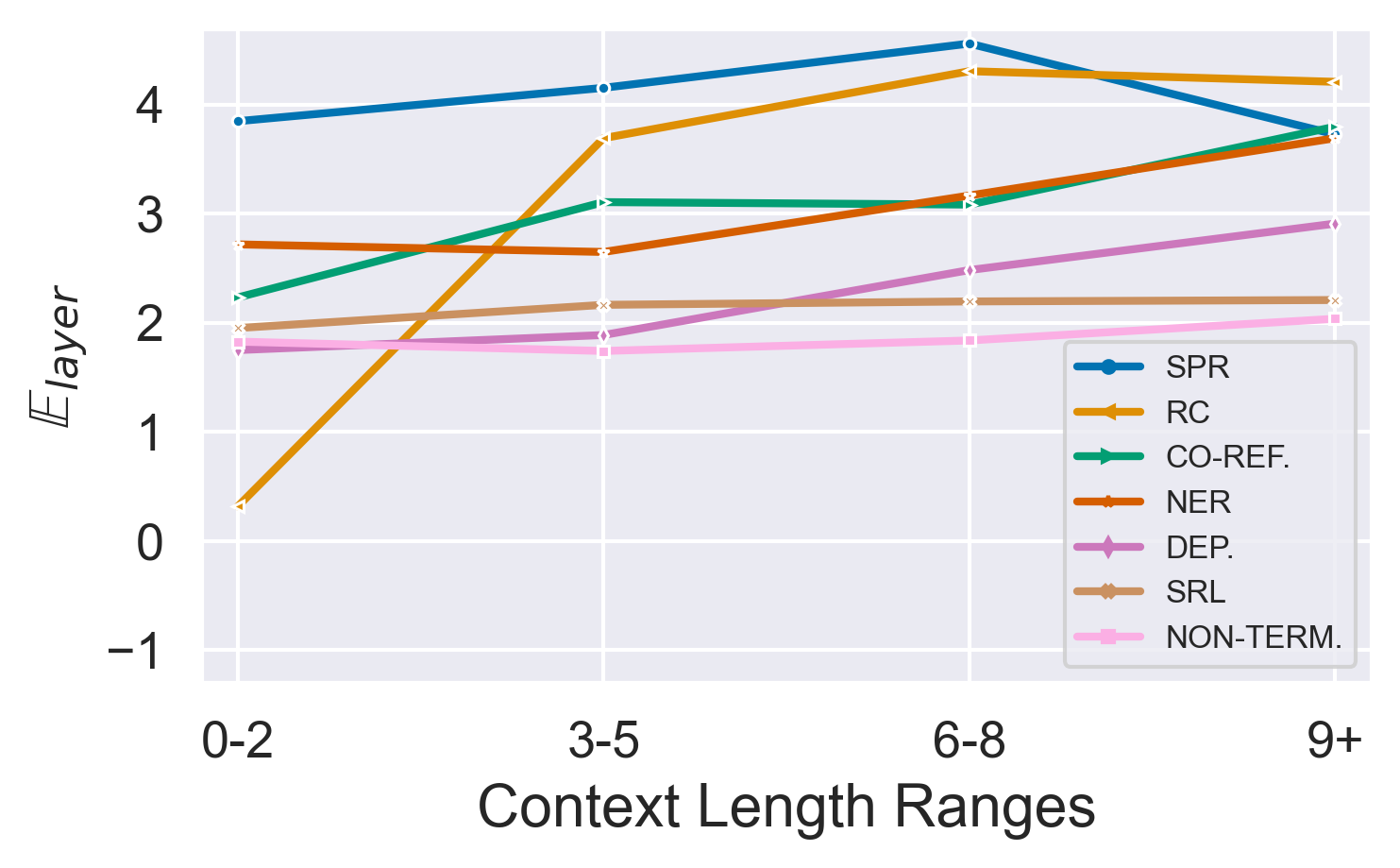}
    \captionsetup{aboveskip=0pt, belowskip=0pt}
    \caption{Expected layers of all seven tasks as a function of context length range.}
    \label{fig:expected_layer_ranges}
\end{figure}

\begin{figure}[h!]
    \includegraphics[width=7cm]{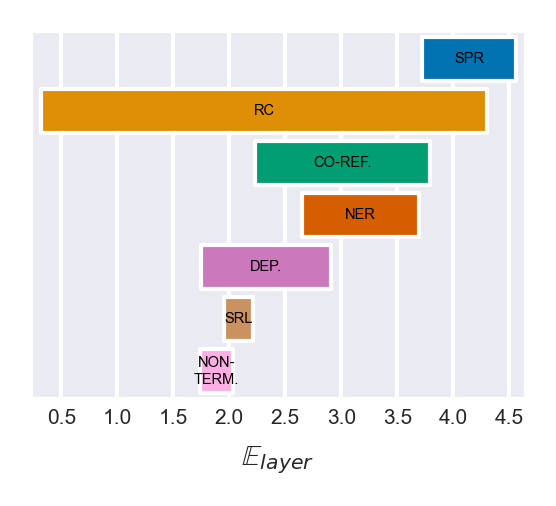}
    \captionsetup{aboveskip=0pt, belowskip=0pt}
    \caption{The range of possible expected layers when varying context length, for each of the seven tasks.}
    \label{fig:expected_layer_bars}
\end{figure}

Our results (Figure \ref{fig:expected_layer_ranges}) allow computing the range of possible expected layers for a task, that may result from taking any context length distribution (Figure \ref{fig:expected_layer_bars}).
The figure shows the wide range of possible relative behaviors of $\el$ for task-pairs: from notable to negligible difference in expected layers (e.g., SRL and Co-ref.), to pairs whose ordering of expected layers may be reversed (i.e., overlapping ranges, such as with SPR and RC). In fact, by taking into account every possible combination of context length distribution for each of the tasks, we get as many as 196 possible rankings of the seven tasks according to their $\el$. One such possible order is, for example, Non-term. < Dep. < SRL < RC < NER < Co-ref. < SPR. We elaborate on this in \S\ref{subsec:Extreme Expected Layer Differences}.

To recap, we find that the difference in $\el$ between some tasks may considerably change and their order may reverse, depending on the context length. This finding lends further support to our claim that mediators should be taken into account. 

\section{Conclusion}

We showed that when performing edge probing to identify what layers are responsible for addressing what tasks, it is imperative to take into account potential mediators, as they may be responsible for much of the observed effect. Specifically, we showed that context length has a significant impact on a task's $\el$. Our analysis shows the wide range of relative orderings of the expected layers for different tasks when assuming different context length distributions; 
from extreme edge cases, like the one we observed in \S\ref{subsubsec:Interesting Case}, to more common, but potentially misleading ones, where the difference between expected layers may dramatically increase or decrease depending on the context length distribution.  Most importantly, it shows that by manipulating the context length distribution, we may get a wide range of outcomes.

Our work suggests that mediating factors should be taken into account when basing analysis on the $\el$. On a broader note, alternative hypotheses should be considered, before limiting oneself to a single interpretation.

Future work will consider the effect of other mediating factors. The two methods we used, NDE and controlled effect, can be used to examine the impact of other mediating factors and should be adopted as part of the field's basic analysis toolkit  \citep[cf.][]{feder2020causalm,vig2020causal}. 
NDE should be used when several effects are examined simultaneously, as it facilitates the assessment of their effect on the tasks' complexity. It is also advisable to use NDE when a more practical examination is required, i.e., when distributions of the mediators are given empirically, as it is easier to derive the mediating factors' impact using this method. In contrast, the controlled effect method should be used when examining the effects of two variables (e.g., tasks and mediating factors) or when comparing several tasks with one mediating effect. 

\section*{Acknowledgements}
This work was supported by the Israel Science Foundation (grant no. 929/17). We would also like to thank Amir Feder for his very insightful feedback on our paper.

\bibliography{references}

\begin{thebibliography}{27}
\expandafter\ifx\csname natexlab\endcsname\relax\def\natexlab#1{#1}\fi

\bibitem[{Belinkov et~al.(2020)Belinkov, Durrani, Dalvi, Sajjad, and
  Glass}]{belinkov-etal-2020-linguistic}
Yonatan Belinkov, Nadir Durrani, Fahim Dalvi, Hassan Sajjad, and James Glass.
  2020.
\newblock \href {https://doi.org/10.1162/coli_a_00367} {On the linguistic
  representational power of neural machine translation models}.
\newblock \emph{Computational Linguistics}, 46(1):1--52.

\bibitem[{Blevins et~al.(2018)Blevins, Levy, and
  Zettlemoyer}]{blevins-etal-2018-deep}
Terra Blevins, Omer Levy, and Luke Zettlemoyer. 2018.
\newblock \href {https://doi.org/10.18653/v1/P18-2003} {Deep {RNN}s encode soft
  hierarchical syntax}.
\newblock In \emph{Proceedings of the 56th Annual Meeting of the Association
  for Computational Linguistics (Volume 2: Short Papers)}, pages 14--19,
  Melbourne, Australia. Association for Computational Linguistics.

\bibitem[{Choshen and Abend(2019)}]{choshen-abend-2019-automatically}
Leshem Choshen and Omri Abend. 2019.
\newblock \href {https://doi.org/10.18653/v1/K19-1028} {Automatically
  extracting challenge sets for non-local phenomena in neural machine
  translation}.
\newblock In \emph{Proceedings of the 23rd Conference on Computational Natural
  Language Learning (CoNLL)}, pages 291--303, Hong Kong, China. Association for
  Computational Linguistics.

\bibitem[{de~Marneffe et~al.(2006)de~Marneffe, MacCartney, and
  Manning}]{de-marneffe-etal-2006-generating}
Marie-Catherine de~Marneffe, Bill MacCartney, and Christopher~D. Manning. 2006.
\newblock \href {http://www.lrec-conf.org/proceedings/lrec2006/pdf/440_pdf.pdf}
  {Generating typed dependency parses from phrase structure parses}.
\newblock In \emph{Proceedings of the Fifth International Conference on
  Language Resources and Evaluation ({LREC}{'}06)}, Genoa, Italy. European
  Language Resources Association (ELRA).

\bibitem[{Devlin et~al.(2019)Devlin, Chang, Lee, and
  Toutanova}]{devlin-etal-2019-bert}
Jacob Devlin, Ming-Wei Chang, Kenton Lee, and Kristina Toutanova. 2019.
\newblock \href {https://doi.org/10.18653/v1/N19-1423} {{BERT}: Pre-training of
  deep bidirectional transformers for language understanding}.
\newblock In \emph{Proceedings of the 2019 Conference of the North {A}merican
  Chapter of the Association for Computational Linguistics: Human Language
  Technologies, Volume 1 (Long and Short Papers)}, pages 4171--4186,
  Minneapolis, Minnesota. Association for Computational Linguistics.

\bibitem[{Elazar et~al.(2021)Elazar, Ravfogel, Jacovi, and
  Goldberg}]{elazar2021amnesic}
Yanai Elazar, Shauli Ravfogel, Alon Jacovi, and Yoav Goldberg. 2021.
\newblock Amnesic probing: Behavioral explanation with amnesic counterfactuals.
\newblock \emph{Transactions of the Association for Computational Linguistics},
  9:160--175.

\bibitem[{Feder et~al.(2020)Feder, Oved, Shalit, and
  Reichart}]{feder2020causalm}
Amir Feder, Nadav Oved, Uri Shalit, and Roi Reichart. 2020.
\newblock \href {http://arxiv.org/abs/2005.13407} {Causalm: Causal model
  explanation through counterfactual language models}.

\bibitem[{Hacohen et~al.(2020)Hacohen, Choshen, and Weinshall}]{hacohen2019all}
Guy Hacohen, Leshem Choshen, and D.~Weinshall. 2020.
\newblock Let's agree to agree: Neural networks share classification order on
  real datasets.
\newblock \emph{International Conference of Machine Learning}.

\bibitem[{Hendrickx et~al.(2009)Hendrickx, Kim, Kozareva, Nakov,
  {\'O}~S{\'e}aghdha, Pad{\'o}, Pennacchiotti, Romano, and
  Szpakowicz}]{hendrickx-etal-2009-semeval}
Iris Hendrickx, Su~Nam Kim, Zornitsa Kozareva, Preslav Nakov, Diarmuid
  {\'O}~S{\'e}aghdha, Sebastian Pad{\'o}, Marco Pennacchiotti, Lorenza Romano,
  and Stan Szpakowicz. 2009.
\newblock \href {https://www.aclweb.org/anthology/W09-2415} {{S}em{E}val-2010
  task 8: Multi-way classification of semantic relations between pairs of
  nominals}.
\newblock In \emph{Proceedings of the Workshop on Semantic Evaluations: Recent
  Achievements and Future Directions ({SEW}-2009)}, pages 94--99, Boulder,
  Colorado. Association for Computational Linguistics.

\bibitem[{Karpathy et~al.(2015)Karpathy, Johnson, and
  Fei-Fei}]{karpathy2015visualizing}
Andrej Karpathy, Justin Johnson, and Li~Fei-Fei. 2015.
\newblock \href {http://arxiv.org/abs/1506.02078} {Visualizing and
  understanding recurrent networks}.

\bibitem[{Kaushik et~al.(2020)Kaushik, Hovy, and Lipton}]{Kaushik2020Learning}
Divyansh Kaushik, Eduard Hovy, and Zachary Lipton. 2020.
\newblock \href {https://openreview.net/forum?id=Sklgs0NFvr} {Learning the
  difference that makes a difference with counterfactually-augmented data}.
\newblock In \emph{International Conference on Learning Representations}.

\bibitem[{Li et~al.(2016)Li, Chen, Hovy, and
  Jurafsky}]{li-etal-2016-visualizing}
Jiwei Li, Xinlei Chen, Eduard Hovy, and Dan Jurafsky. 2016.
\newblock \href {https://doi.org/10.18653/v1/N16-1082} {Visualizing and
  understanding neural models in {NLP}}.
\newblock In \emph{Proceedings of the 2016 Conference of the North {A}merican
  Chapter of the Association for Computational Linguistics: Human Language
  Technologies}, pages 681--691, San Diego, California. Association for
  Computational Linguistics.

\bibitem[{Li et~al.(2015)Li, Yosinski, Clune, Lipson, and
  Hopcroft}]{Li2015ConvergentLD}
Y.~Li, J.~Yosinski, J.~Clune, H.~Lipson, and J.~Hopcroft. 2015.
\newblock Convergent learning: Do different neural networks learn the same
  representations?
\newblock In \emph{FE@NIPS}.

\bibitem[{Pearl(2001)}]{10.5555/647235.720084}
Judea Pearl. 2001.
\newblock Direct and indirect effects.
\newblock In \emph{Proceedings of the 17th Conference in Uncertainty in
  Artificial Intelligence}, UAI '01, page 411–420, San Francisco, CA, USA.
  Morgan Kaufmann Publishers Inc.

\bibitem[{Peters et~al.(2018)Peters, Neumann, Zettlemoyer, and
  Yih}]{peters-etal-2018-dissecting}
Matthew Peters, Mark Neumann, Luke Zettlemoyer, and Wen-tau Yih. 2018.
\newblock \href {https://doi.org/10.18653/v1/D18-1179} {Dissecting contextual
  word embeddings: Architecture and representation}.
\newblock In \emph{Proceedings of the 2018 Conference on Empirical Methods in
  Natural Language Processing}, pages 1499--1509, Brussels, Belgium.
  Association for Computational Linguistics.

\bibitem[{Reisinger et~al.(2015)Reisinger, Rudinger, Ferraro, Harman, Rawlins,
  and Van~Durme}]{reisinger-etal-2015-semantic}
Drew Reisinger, Rachel Rudinger, Francis Ferraro, Craig Harman, Kyle Rawlins,
  and Benjamin Van~Durme. 2015.
\newblock \href {https://doi.org/10.1162/tacl_a_00152} {Semantic proto-roles}.
\newblock \emph{Transactions of the Association for Computational Linguistics},
  3:475--488.

\bibitem[{Rogers et~al.(2020)Rogers, Kovaleva, and
  Rumshisky}]{rogers2020primer}
Anna Rogers, Olga Kovaleva, and Anna Rumshisky. 2020.
\newblock \href {http://arxiv.org/abs/2002.12327} {A primer in bertology: What
  we know about how bert works}.

\bibitem[{Schlichtkrull et~al.(2020)Schlichtkrull, Cao, and
  Titov}]{Schlichtkrull2020InterpretingGN}
M.~Schlichtkrull, Nicola~De Cao, and Ivan Titov. 2020.
\newblock Interpreting graph neural networks for nlp with differentiable edge
  masking.

\bibitem[{Sennrich(2017)}]{sennrich-2017-grammatical}
Rico Sennrich. 2017.
\newblock \href {https://www.aclweb.org/anthology/E17-2060} {How grammatical is
  character-level neural machine translation? assessing {MT} quality with
  contrastive translation pairs}.
\newblock In \emph{Proceedings of the 15th Conference of the {E}uropean Chapter
  of the Association for Computational Linguistics: Volume 2, Short Papers},
  pages 376--382, Valencia, Spain. Association for Computational Linguistics.

\bibitem[{Silveira et~al.(2014)Silveira, Dozat, de~Marneffe, Bowman, Connor,
  Bauer, and Manning}]{silveira-etal-2014-gold}
Natalia Silveira, Timothy Dozat, Marie-Catherine de~Marneffe, Samuel Bowman,
  Miriam Connor, John Bauer, and Chris Manning. 2014.
\newblock \href
  {http://www.lrec-conf.org/proceedings/lrec2014/pdf/1089_Paper.pdf} {A gold
  standard dependency corpus for {E}nglish}.
\newblock In \emph{Proceedings of the Ninth International Conference on
  Language Resources and Evaluation ({LREC}'14)}, pages 2897--2904, Reykjavik,
  Iceland. European Language Resources Association (ELRA).

\bibitem[{Tenney et~al.(2019{\natexlab{a}})Tenney, Das, and
  Pavlick}]{tenney-etal-2019-bert}
Ian Tenney, Dipanjan Das, and Ellie Pavlick. 2019{\natexlab{a}}.
\newblock \href {https://doi.org/10.18653/v1/P19-1452} {{BERT} rediscovers the
  classical {NLP} pipeline}.
\newblock In \emph{Proceedings of the 57th Annual Meeting of the Association
  for Computational Linguistics}, pages 4593--4601, Florence, Italy.
  Association for Computational Linguistics.

\bibitem[{Tenney et~al.(2019{\natexlab{b}})Tenney, Xia, Chen, Wang, Poliak,
  McCoy, Kim, Durme, Bowman, Das, and Pavlick}]{tenney2018what}
Ian Tenney, Patrick Xia, Berlin Chen, Alex Wang, Adam Poliak, R~Thomas McCoy,
  Najoung Kim, Benjamin~Van Durme, Sam Bowman, Dipanjan Das, and Ellie Pavlick.
  2019{\natexlab{b}}.
\newblock \href {https://openreview.net/forum?id=SJzSgnRcKX} {What do you learn
  from context? probing for sentence structure in contextualized word
  representations}.
\newblock In \emph{International Conference on Learning Representations}.

\bibitem[{Vig et~al.(2020)Vig, Gehrmann, Belinkov, Qian, Nevo, Singer, and
  Shieber}]{vig2020causal}
Jesse Vig, Sebastian Gehrmann, Yonatan Belinkov, Sharon Qian, Daniel Nevo,
  Yaron Singer, and Stuart Shieber. 2020.
\newblock \href {http://arxiv.org/abs/2004.12265} {Causal mediation analysis
  for interpreting neural nlp: The case of gender bias}.

\bibitem[{Wang et~al.(2019)Wang, Tenney, Pruksachatkun, Yu, Hula, Xia,
  Pappagari, Jin, McCoy, Patel, Huang, Phang, Grave, Kim, Htut, F'{e}vry, Chen,
  Nangia, Liu, , Mohananey, Bordia, Pavlick, and Bowman}]{wang2019jiant}
Alex Wang, Ian~F. Tenney, Yada Pruksachatkun, Katherin Yu, Jan Hula, Patrick
  Xia, Raghu Pappagari, Shuning Jin, R.~Thomas McCoy, Roma Patel, Yinghui
  Huang, Jason Phang, Edouard Grave, Najoung Kim, Phu~Mon Htut, Thibault
  F'{e}vry, Berlin Chen, Nikita Nangia, Haokun Liu, , Anhad Mohananey, Shikha
  Bordia, Ellie Pavlick, and Samuel~R. Bowman. 2019.
\newblock {jiant} 1.0: A software toolkit for research on general-purpose text
  understanding models.
\newblock \url{http://jiant.info/}.

\bibitem[{Weischedel et~al.(2013)Weischedel, Palmer, Marcus, Hovy, Pradhan,
  Ramshaw, Xue, Taylor, Kaufman, Franchini et~al.}]{weischedel2013ontonotes}
Ralph Weischedel, Martha Palmer, Mitchell Marcus, Eduard Hovy, Sameer Pradhan,
  Lance Ramshaw, Nianwen Xue, Ann Taylor, Jeff Kaufman, Michelle Franchini,
  et~al. 2013.
\newblock Ontonotes release 5.0 ldc2013t19.
\newblock \emph{Linguistic Data Consortium, Philadelphia, PA}, 23.

\bibitem[{Xu et~al.(2009)Xu, Kang, Ringgaard, and
  Och}]{10.5555/1620754.1620790}
Peng Xu, Jaeho Kang, Michael Ringgaard, and Franz Och. 2009.
\newblock Using a dependency parser to improve smt for subject-object-verb
  languages.
\newblock In \emph{Proceedings of Human Language Technologies: The 2009 Annual
  Conference of the North American Chapter of the Association for Computational
  Linguistics}, NAACL '09, page 245–253, USA. Association for Computational
  Linguistics.

\bibitem[{Yosinski et~al.(2015)Yosinski, Clune, Nguyen, Fuchs, and
  Lipson}]{DBLP:journals/corr/YosinskiCNFL15}
Jason Yosinski, Jeff Clune, Anh~Mai Nguyen, Thomas~J. Fuchs, and Hod Lipson.
  2015.
\newblock \href {http://arxiv.org/abs/1506.06579} {Understanding neural
  networks through deep visualization}.
\newblock \emph{CoRR}, abs/1506.06579.

\end{thebibliography}
\bibliographystyle{acl_natbib}
\clearpage
\pdfoutput=1
\appendix
\section{Appendix}\label{sec:appendix}
    \subsection{Additional Example of the Extreme Case}\label{subsec:Additional Interesting Case}
\begin{figure}[h!]
            \includegraphics[width=7cm]{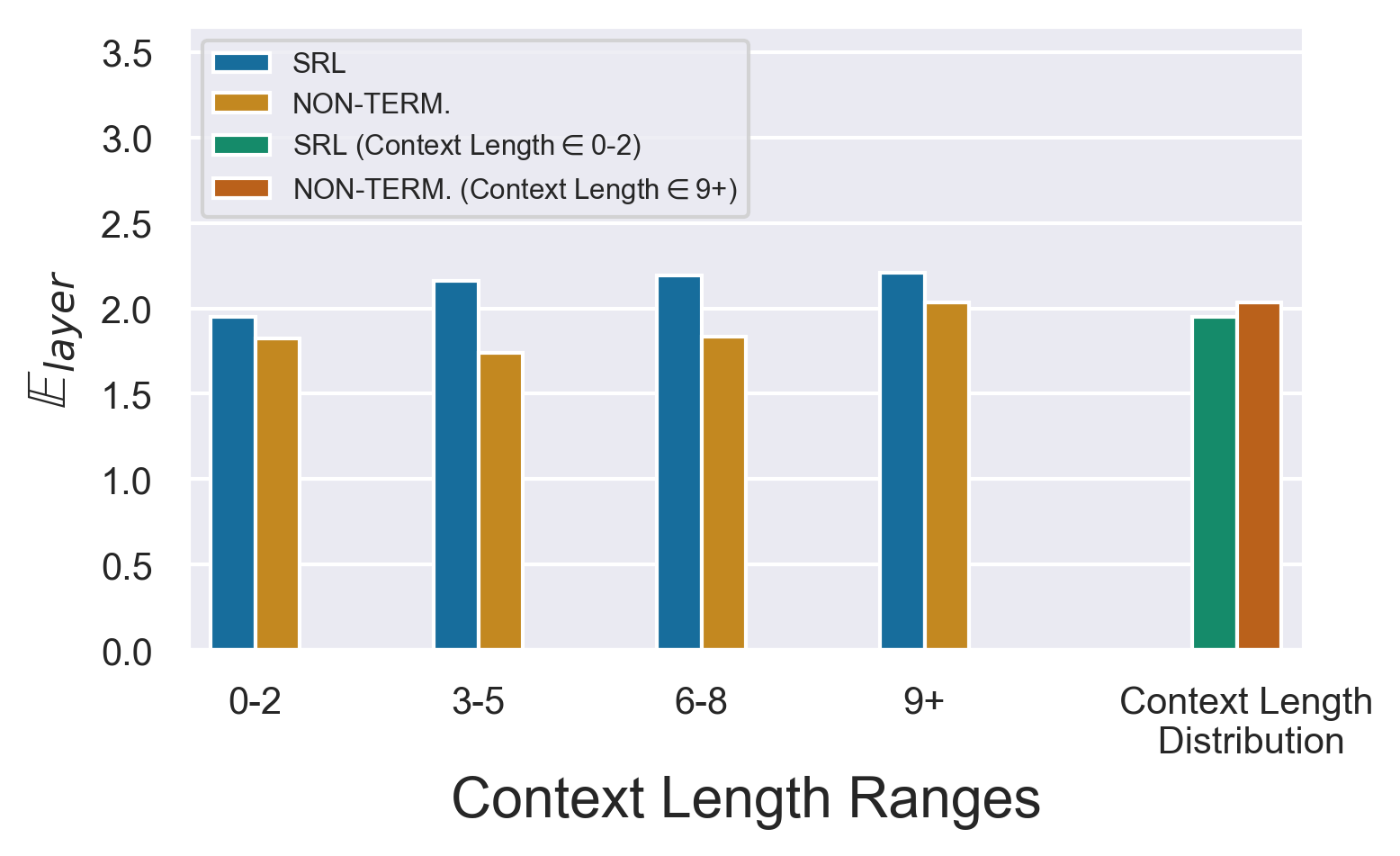}
            \captionsetup{aboveskip=0pt, belowskip=0pt}
            \caption{$\el$ of SRL and Non-term. for different context length ranges (4 left blue and yellow pairs), and their $\el$ when all instances of SRL are of context length $l\in [0,2]$ and all those of Non-term. are of context length $l\geq 9$ (rightmost green and red pair). While for every context length range, SRL's $\el$ is bigger than that of Non-term., for some context length distribution that order may be reversed.}
            \label{fig:sympson_paradox_2}
        \end{figure}

        We show another example of a task-pair that, under certain distributions of context lengths, exhibits similar behavior to that observed in the edge case described in \S\ref{subsubsec:Interesting Case} (figure \ref{fig:sympson_paradox_2}).
    \subsection{Context Length Distribution}\label{Context Length Distribution}
\begin{figure}[h]
        \includegraphics[width=7cm]{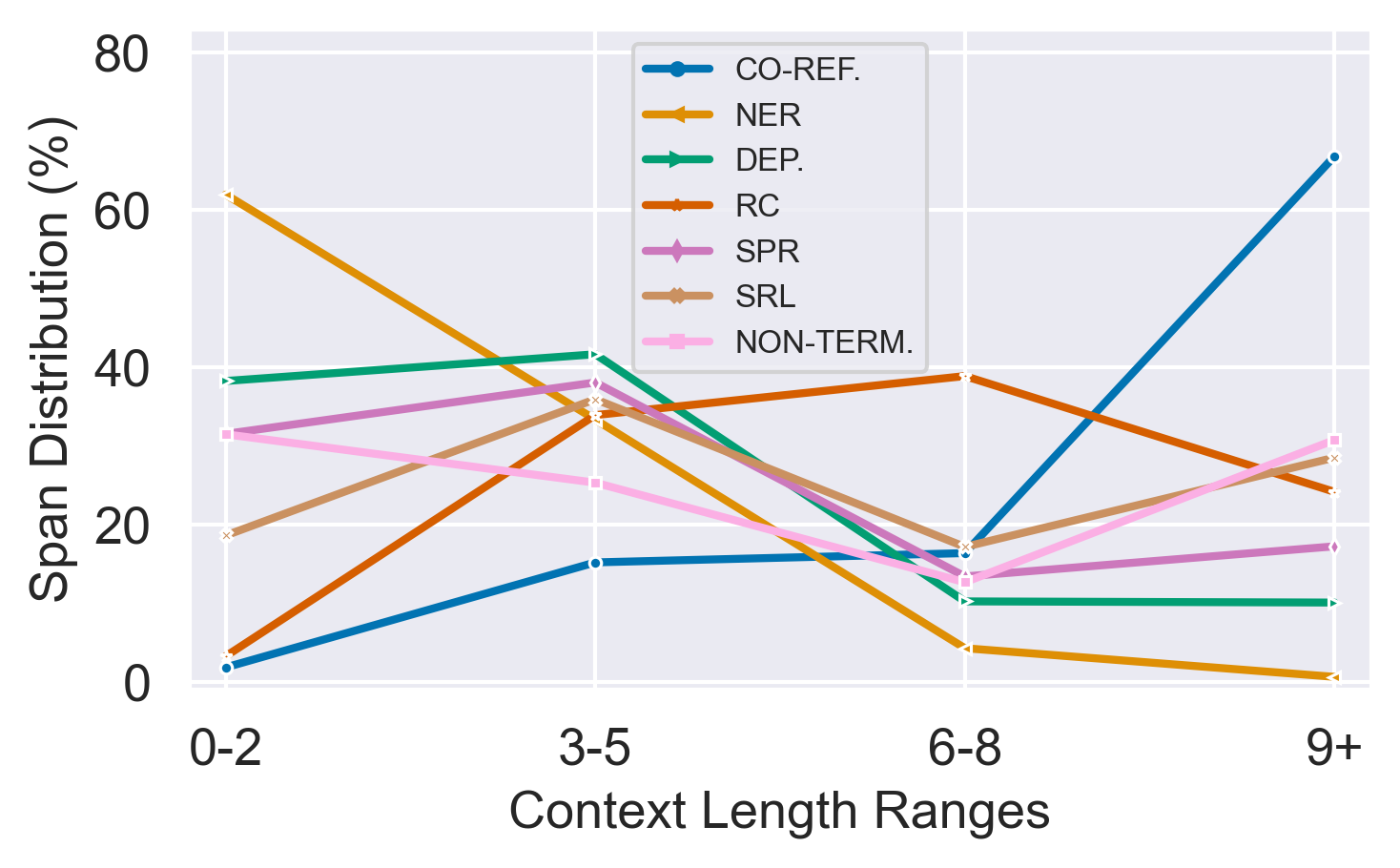}
        \captionsetup{aboveskip=0pt, belowskip=0pt}
        \caption{Percentage of examples as a function of context length range, for each of the 7 tasks (see legend).}
        \label{fig:probability_ranges}
    \end{figure}

    A lot of our work deals with possible context length distributions, normalizing distribution, and accounting for the distribution. We provide here the actual distributions which are the underlying property controlling the seen effects.
    We provide data on the percentage of examples in each context length range for each task (figure \ref{fig:probability_ranges}).
    
    \subsection{NDE vs. Unmediated Difference for All Task-Pairs}\label{subsec:NDE vs. Unmediated difference for All Task-pairs}
\begin{figure*}[h]
        \includegraphics[width=15cm]{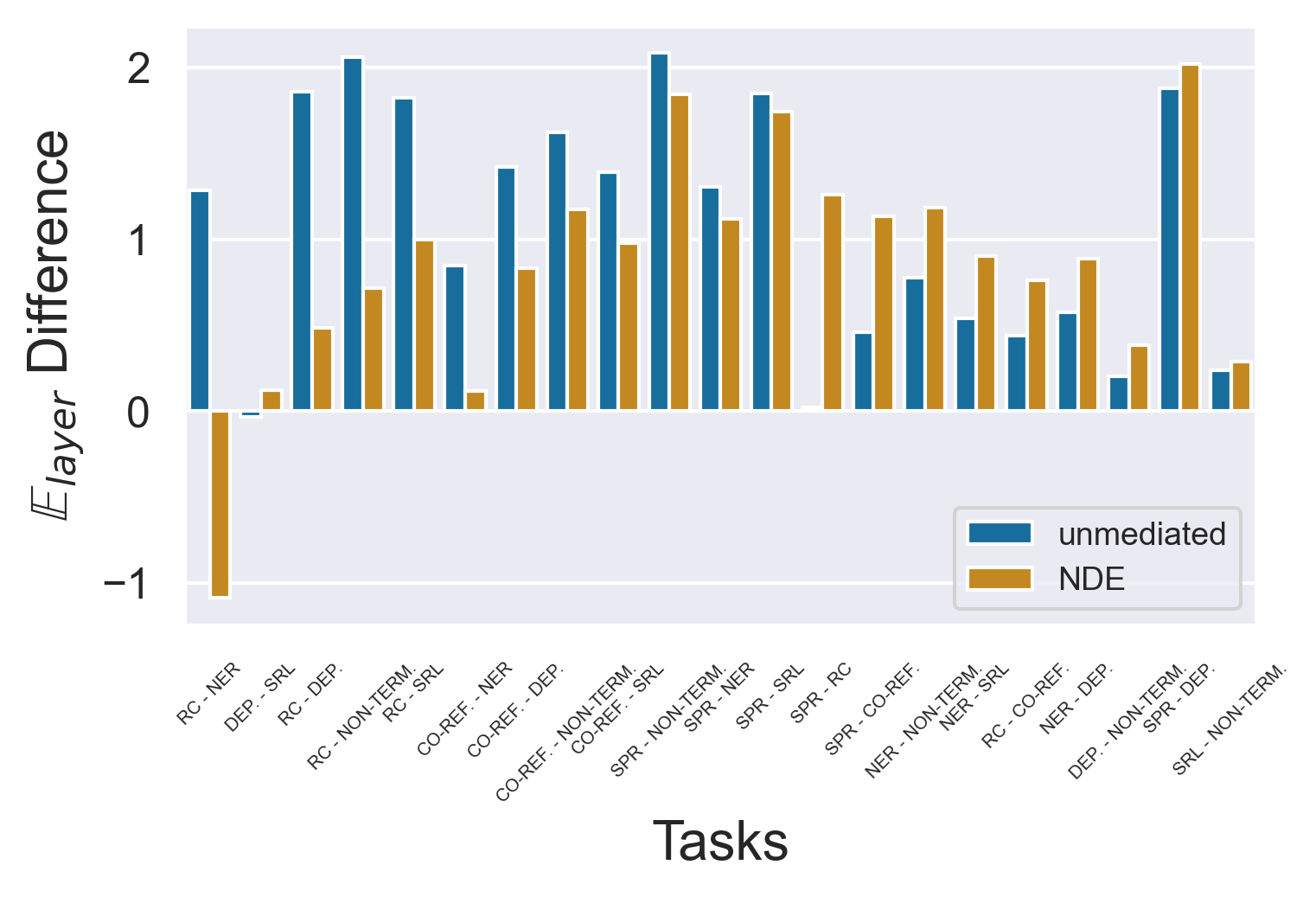}
        \captionsetup{aboveskip=0pt, belowskip=0pt}
        \caption{Difference between unmediated $\el{}$ and NDE for every task-pair. The employed context length distributions (as part of the NDE calculations) are, from left to right, of NER, SRL, Dep., Non-term., SRL, Co-ref., Dep., Non-term., SRL, Non-term., SPR, SRL, SPR, SPR, Non-term., SRL, RC, NER, Non-term., Dep. and SRL.}
        \label{fig:NDE vs. Unmediated difference for All Task-pairs}
    \end{figure*}

    For every task-pair, we compare the unmediated $\el$ difference with the pair's NDE. Figure \ref{fig:NDE vs. Unmediated difference for All Task-pairs} presents this comparison for each task-pair, with the distribution of one of the pair's tasks being applied in the NDE calculations, for each task-pair.

    \subsection{Extreme $\el$ Differences}\label{subsec:Extreme Expected Layer Differences}
\begin{figure*}[h]
        \includegraphics[width=18cm]{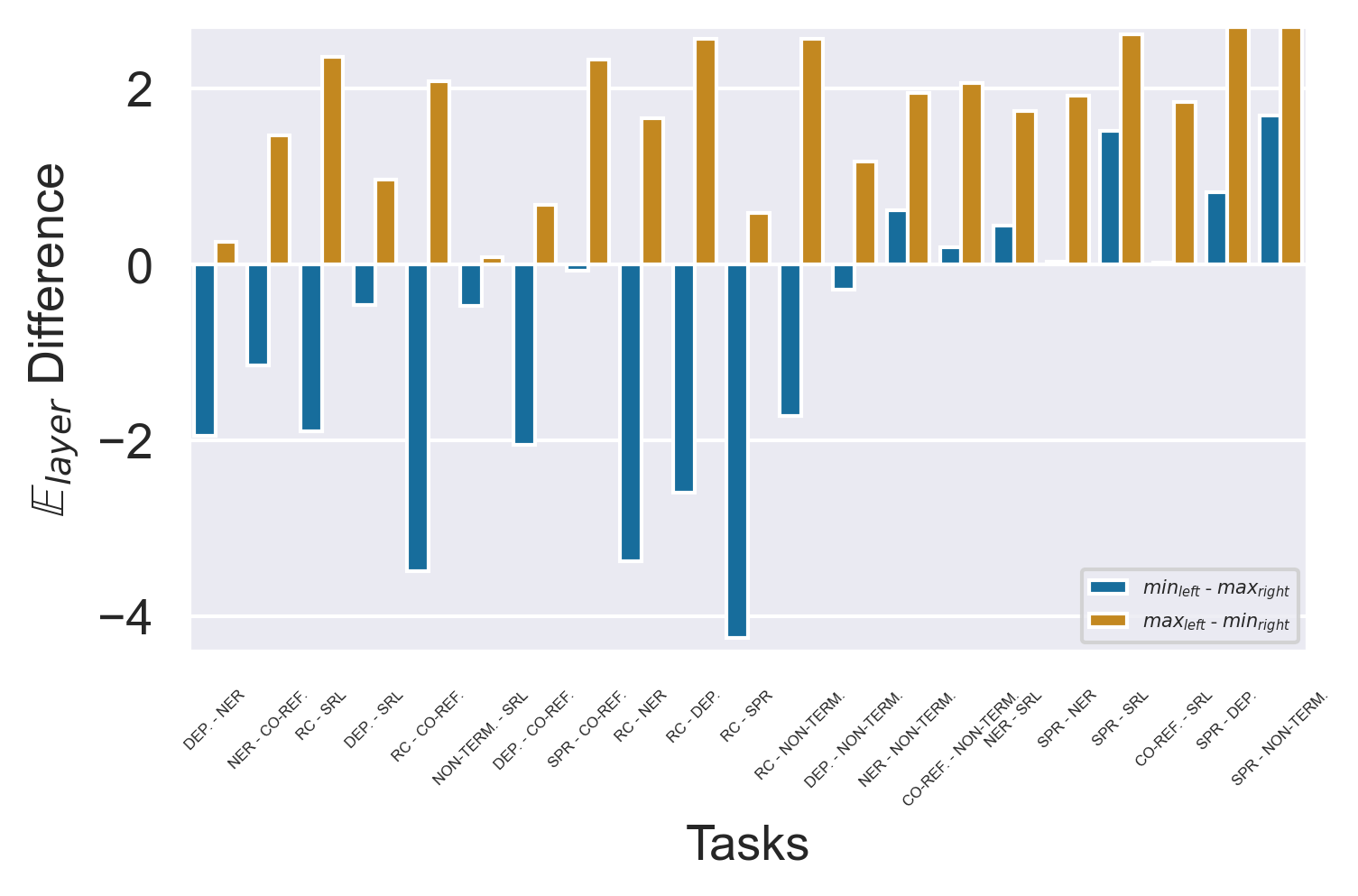}
        \captionsetup{aboveskip=0pt, belowskip=0pt}
        \caption{Difference between the minimal possible expected layer of the left task and the maximal possible expected layer of the right task (blue - see legend), and vice-versa (yellow - see legend), for every task-pair.}
        \label{fig:max_min}
    \end{figure*}

    Based on figure \ref{fig:expected_layer_bars}, we compute the extreme $\el$ differences of each task-pair. Namely, for each such pair, we juxtapose the difference between the maximal possible $\el$ of the first task and the minimal $\el$ of the second one with the opposite case (the difference between the minimal possible $\el$ of the first task and the maximal $\el$ of the second one). Our results can be seen in figure \ref{fig:max_min}.

\end{document}